\documentclass[letterpaper]{article} 
\usepackage{aaai2026}  
\usepackage{times}  
\usepackage{helvet}  
\usepackage{courier}  
\usepackage[hyphens]{url}  
\usepackage{graphicx} 
\usepackage{natbib}  
\usepackage{caption} 
\usepackage{booktabs}
\usepackage{amsmath}
\usepackage{amsfonts}
\usepackage{bm}
\usepackage{makecell} 
\usepackage{algorithm}
\usepackage{algorithmic}

\usepackage{newfloat}
\usepackage{listings}
\DeclareCaptionStyle{ruled}{labelfont=normalfont,labelsep=colon,strut=off} 
\floatstyle{ruled}
\newfloat{listing}{tb}{lst}{}
\floatname{listing}{Listing}
\title{SGMAGNet: A Baseline Model for 3D Cloud Phase Structure Reconstruction on a New Passive–Active Satellite Benchmark}

 \author{
     Chi Yang\textsuperscript{\rm 1}\equalcontrib,
     Fu Wang\thanks{(This work is supported by National Natural Science Foundation of China under Grant 42471437, corresponding author: Fu Wang, Wangfu@cma.cn}\textsuperscript{\rm 2,3}\equalcontrib,
     Xiaofei Yang\textsuperscript{\rm 4},\equalcontrib,
     Hao Huang\textsuperscript{\rm 3},
     Weijia Cao\textsuperscript{\rm 5},
     Xiaowen Chu\textsuperscript{\rm 6}
 }
 \affiliations{
     \textsuperscript{\rm 1}the School of Computer Science and Engineering, Sichuan University of Science and Engineering\\
     \textsuperscript{\rm 2}CMA Earth System Modeling and Predictiopn Centre (CEMC)\\
     \textsuperscript{\rm 3}State Key Laboratory of Severe Weather Meteorological Science and Technology(LaSW)\\
     No.46 Zhongguancun North Street, Beijing, China\\    
     \textsuperscript{\rm 4}the School of Electronics and Communication Engineering, Guangzhou University\\
     \textsuperscript{\rm 5}Aerospace Information Research Institute, Chinese Academy of Sciences\\
     \textsuperscript{\rm 6}The Hong Kong University of Science and Technology (Guangzhou)\\
}

\usepackage{bibentry}

\begin{document}

\maketitle

\begin{abstract}%
Cloud phase profiles are critical for numerical weather prediction (NWP), as they directly affect radiative transfer and precipitation processes. In this study, we present a benchmark dataset and a baseline framework for transforming multimodal satellite observations into detailed 3D cloud phase structures, aiming toward operational cloud phase profile retrieval and future integration with NWP systems to improve cloud microphysics parameterization. The multimodal observations consist of (1) high–spatiotemporal–resolution, multi-band visible (VIS) and thermal infrared (TIR) imagery from geostationary satellites, and (2) accurate vertical cloud phase profiles from spaceborne lidar (CALIOP/CALIPSO) and radar (CPR/CloudSat). The dataset consists of synchronized image–profile pairs across diverse cloud regimes, defining a supervised learning task: given VIS/TIR patches, predict the corresponding 3D cloud phase structure. We adopt \textit{SGMAGNet} as the main model and compare it with several baseline architectures, including UNet variants and \textit{SegNet}, all designed to capture multi-scale spatial patterns. Model performance is evaluated using standard classification metrics, including \textit{Precision}, \textit{Recall}, \textit{F1-score}, and \textit{IoU}. The results demonstrate that \textit{SGMAGNet} achieves superior performance in cloud phase reconstruction, particularly in complex multi-layer and boundary transition regions. Quantitatively, \textit{SGMAGNet} attains a \textit{Precision} of 0.922, \textit{Recall} of 0.858, \textit{F1-score} of 0.763, and an \textit{IoU} of 0.617, significantly outperforming all baselines across these key metrics.

\end{abstract}%


\section{Introduction}
Clouds play a pivotal role in regulating the Earth's climate system, significantly influencing the global energy balance by reflecting solar radiation and absorbing longwave radiation~\cite{norris2016evidence}. The three-dimensional (3D) structure of clouds is crucial for numerical weather prediction and climate change studies. Specifically, the vertical layering of different cloud types—comprising ice, water, and mixed-phase clouds—plays a decisive role in the accurate numerical simulation of physical processes. The structural characteristics of these clouds directly impact key climate factors, such as radiation transport, precipitation processes, and atmospheric stability~\cite{jeggle2024icecloudnet}. Therefore, precise modeling of the 3D distribution of clouds is essential for improving weather forecasting accuracy and advancing our understanding of climate change mechanisms.

In this context, active lidar and cloud radar observations provide high-precision 3D structural information about clouds and can distinguish various cloud phases, such as ice and water clouds~\cite{lin2025clann}. Lidar is particularly sensitive to cloud tops, while cloud radar is more responsive to cloud bottoms. Although lidar can more accurately reflect the structural characteristics of multilayered clouds, satellite-based active remote sensing data remains relatively scarce. Furthermore, these observations are often limited to a single contour and require long-term data accumulation to reveal the climatic patterns of global cloud structures~\cite{barker20113d}.

Geostationary satellite multi-channel imagers offer an alternative approach for obtaining the 3D structure of clouds. These instruments provide high temporal resolution measurements of the vertical distribution of water vapor, enabling the retrieval of cloud top height, multilayer cloud characteristics, and cloud type information. This yields a rich set of data on the 3D structural properties of clouds~\cite{foley20243d, ham2015improving}. As multiple geostationary satellite imagers around the world share similar functionalities and characteristics, they have the potential to provide comprehensive global observations with high temporal resolution.

Recent studies have attempted to combine active and passive observations in a multimodal approach, exploring the feasibility of transferring 3D cloud vertical structure features from active observations to passive multi-channel imagers~\cite{lin2025clann, guillaume2018horizontal}. However, a comprehensive thematic dataset that fully captures the vertical structure of various cloud types remains absent.

\subsection{Problem Formulation}
The aim of this study is to invert the 3D cloud structure of a target region based on 2D multi-channel observations provided by a geostationary imager (e.g., AHI). Specifically, the input is a 2D remote sensing image with high temporal resolution and multispectral channels, and the goal is to reconstruct the cloud vertical profile information, i.e., the 3D structure of the cloud, in the corresponding region~\cite{wang2019multilayer}. The primary challenge is that geostationary imagers provide only 2D radiometric features and lack direct vertical structure information, while the true 3D cloud structure is sparsely available from polar orbiting satellites, such as CloudSat, equipped with LiDAR, which provide data only along specific scanning paths~\cite{ham2015improving}. As a result, the training data consist of true cloud contour labels for only a limited number of pixels in the AHI images, leading to significant label sparsity. To map 2D observations to 3D structures, it is necessary to establish the relationship between the multi-channel features of the imager and the vertical cloud structures observed by LiDAR. In the training phase, a supervised model is constructed using sparse samples along the LiDAR path~\cite{wang2023retrieving}. In the inference phase, this mapping relationship is generalized to the entire 2D observation plane to predict the 3D cloud structure for all pixel points.

\begin{itemize}
\item Extended the definition of cloud vertical structure and classification tasks to better support applications in areas such as numerical weather prediction and climate change.
\item A benchmark data set based on passive and active observations was constructed.
\item A baseline algorithm for achieving state-of-the-art (SOTA) with the current dataset is proposed.
\end{itemize}

\begin{figure*}[ht] 
\centering 
\includegraphics[scale=0.2]{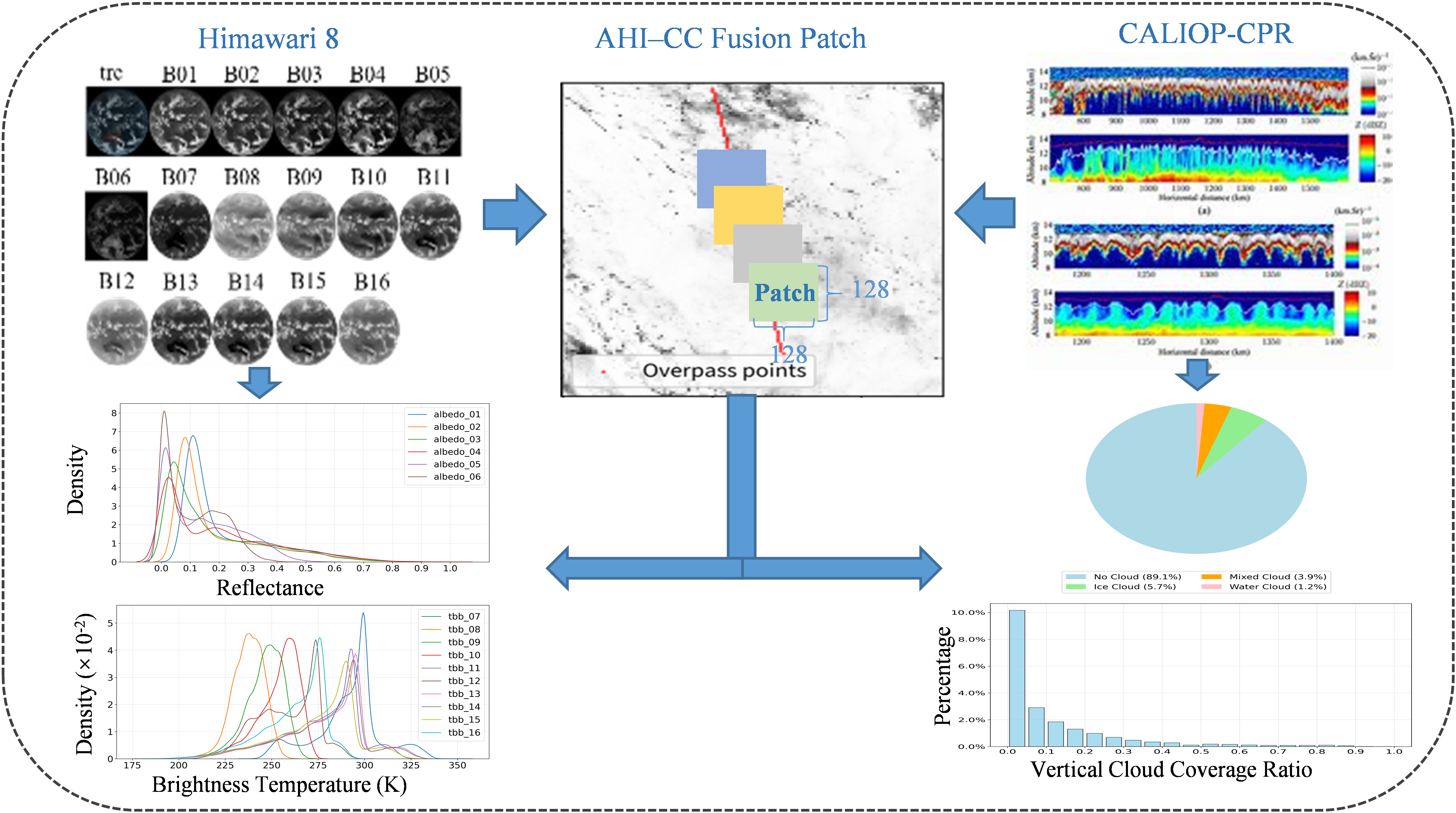} 
\caption{Spatiotemporal Patch Construction for Multi-Modal Cloud Phase Profiles via AHI/Himawari-8 and CALIOP-CPR Fusion. This figure illustrates the fusion process of AHI/Himawari-8 multi-band imagery (B01–B16) with CALIOP-CPR cloud profiling data to construct spatiotemporal patches. The workflow includes extracting 128×128 pixel patches from aligned overpass points, integrating reflectance, brightness temperature, and cloud phase data. Statistical outputs depict density distributions of reflectance and brightness temperature, alongside vertical cloud coverage ratios, highlighting dominant cloud types (e.g., 89.3\% no cloud, 5.7\% ice cloud, 1.2\% water cloud, and 3.9\% mixed cloud  ).} 
\label{fig1} 
\end{figure*}

\section{Related Work}
Traditional observation methods have significant limitations: passive optical sensors are hindered by optical penetrability and cannot directly capture vertical structures such as cloud top height and stratification type~\cite{noh2022framework}. Active radar, while capable of resolving 3D structures, faces issues like orbital sparsity and attenuation in thick clouds, leading to insufficient global coverage and limiting its use for high spatial and temporal resolution cloud field data in climate research~\cite{barker20113d}. Existing studies typically focus on local cloud mask generation or single-height layer statistics, lacking the capability to jointly model vertical stratification across multiple cloud types in full space.

To address these gaps, this study proposes a physical-data dual-driven 3D cloud classification framework. This framework constructs a spatio-temporally matched AHI-CC dataset by fusing AHI multi-channel radiance and CloudSat/CALIPSO vertical profile data. It also introduces a physical loss function with geometric continuity constraints and spatial distribution features of cloud types, alongside a self-supervised pre-training strategy. The goal is to overcome the vertical stratification challenge of hybrid clouds and accurately classify 3D cloud types.

Classical UNet employs an encoder-decoder architecture that extracts high-level features through downsampling and restores spatial details through upsampling, making it particularly effective for image segmentation tasks~\cite{ronneberger2015u}.
SegNet features a lightweight encoder-decoder architecture that excels in two-dimensional segmentation tasks, offering low computational costs and strong boundary localization capabilities, making it particularly well-suited for scenarios with high real-time requirements and limited memory ~\cite{badrinarayanan2017segnet}.
UnetCoNext adopts a novel architectural design with improved skip connections to enhance feature transmission efficiency and improve the clarity of segmentation results~\cite{jeggle2024icecloudnet}.
Attention-based UNet introduces an attention gate mechanism that enables the network to focus on key regions, thereby enhancing segmentation accuracy in complex scenes~\cite{oktay2018attention}.

Transformer was originally used in natural language processing, relying on self-attention mechanisms to model global dependencies. Its advantages have been introduced into image segmentation to compensate for the shortcomings of CNNs in modeling long-range dependencies. TransUNet is one of the representative models, combining the local feature extraction of CNNs with the global modeling capabilities of ViT. It introduces Transformer modules into the encoder to achieve the fusion of local and global information, thereby improving segmentation performance in complex scenarios such as medical and remote sensing images~\cite{chen2021transunet}.

Mamba is an emerging state space model that offers the advantages of efficient modeling of long-range dependencies and linear computational complexity, making it particularly suitable for large-scale sequence data processing. Mamba-UNet incorporates it into the UNet encoder, replacing or enhancing the convolution module, thereby reducing computational costs while improving multi-scale feature modeling capabilities. It performs particularly well in segmentation tasks involving complex texture structures, effectively integrating local details with global information~\cite{wang2024mamba}.

Sgmagnet effectively addresses the limitations of traditional segmentation networks in two-dimensional to three-dimensional mapping and multi-level cloud structure representation by introducing a dynamic encoding mechanism and multi-scale generation module, demonstrating its innovative design in spatial reconstruction and structural inference tasks.


3D cloud structure inversion is crucial for atmospheric remote sensing and climate modeling. Early studies linked mid-latitude cloud decline with greenhouse gas increases but were limited by CloudSat/CALIPSO's sparse coverage and radar signal attenuation. To overcome this, statistical fusion approaches emerged: Barker et al. developed a 3D cloud algorithm for EarthCARE combining passive and active sensors, while Miller et al. extended CloudSat profiles to MODIS for improved coverage~\cite{miller2014estimating}.

Recent advances leverage deep learning and multi-angle data. Forster et al. proposed the "hidden core" concept to simplify 3D radiative transfer in convective clouds~\cite{forster2021toward}. Wang et al. used CGANs to retrieve vertical reflectivity from MODIS~\cite{wang2023retrieving}, Hu et al. introduced CDUNet for enhanced cloud detection from geostationary satellites~\cite{hu2021cdunet}, and Girtsou et al. applied self-supervised learning to improve 3D reconstruction of tropical convection using CloudSat~\cite{girtsou20253d}.

In 2024, physical constraints and global validation took focus. Wang et al. proposed the "CloudMask Loss" function to improve multilayer cloud mask reconstruction~\cite{wang2023toward}, and Miller’s team integrated CloudSat/CALIPSO into NCAR's MET toolkit to quantify model radiation errors~\cite{miller2014model}. Ham et al. enhanced shortwave radiation modeling by expanding 2D cloud profiles into 3D structures~\cite{ham2015improving}.

Currently, research emphasizes multi-modal fusion and operational applications. Lin et al. proposed the CLANN model for 3D cloud estimation from geostationary satellite data, while Noh’s team optimized cloud height retrievals with VIIRS and aerial feedback . Jeggle et al. created IceCloudNet for high-resolution cloud content inversion~\cite{jeggle2024icecloudnet}, and Zhao et al. used Res-UNet to improve global cloud layer accuracy~\cite{bruning2024artificial}. 

\begin{figure*}[ht] 
\centering 
\includegraphics[scale=0.98]{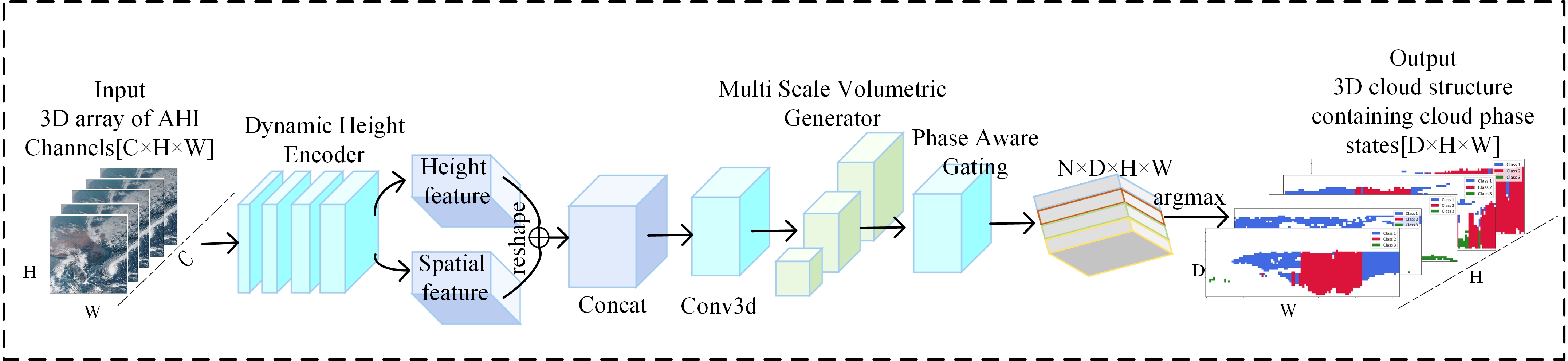} 
\caption{SGMAGNet Framework for Inferring Cloud Phase Profiles from Multispectral Satellite Data. This figure depicts the SGMAGNet architecture, which processes a 2D array of AHI multispectral channels [C×H×W] using a Dynamic Height Encoder to extract spatial features.} 
\label{fig1} 
\end{figure*}

\section{Benchmark Dataset}
This section describes the dataset sources and the principle of spatio-temporal matching. As illustrated in Figure 1, we first construct the dataset by integrating multispectral imagery from the Himawari-8 Advanced Himawari Imager (AHI) and vertical cloud structure data from CALIOP and CPR sensors. The central panel shows the spatio-temporal matching process, where AHI data patches are aligned with CALIOP-CPR overpass points to form AHI–CC fusion samples.

To further characterize the dataset, we compute the probability density distributions of reflectance and brightness temperature for all spectral channels of the imager across matched patches (bottom-left). In addition, the cloud type distribution and vertical cloud coverage ratio derived from the CALIOP-CPR observations are visualized (bottom-right), highlighting the sparsity and diversity of cloud profiles. These statistical summaries provide insight into the spectral and structural characteristics of the constructed dataset.

\subsection{Satellite Data}

Imageries from AHI/Himawari-8,
The AHI-8 L1 dataset is provided by the Japanese meteorological satellite Himawari-8 and primarily consists of radiance data from the Advanced High Resolution Imager (AHI), which is used for meteorological monitoring and environmental analysis. This dataset offers high spatial resolution (2 km) and a temporal resolution of 10 minutes. Data links: \url{https://www.eorc.jaxa.jp/ptree/userguide.html.}
Joint Prodcut cloud product of lidar and radar, CALIPSO-CPR data setProvided by NASA, this dataset combines LIDAR technology with other sensor data for cloud classification and atmospheric structure analysis. It offers detailed spatial and temporal information on cloud distribution, layers, and other characteristics. Data link: \url{http://www.cloudsat.cira.colostate.edu/data-Products at level 2b.}

\subsection{Matched Dataset}
In this study, we focus on the period from October to December 2018, covering the region between 100°–160°E longitude and 60°S–60°N latitude. The input features include 16 spectral channels from the AHI-8 L1 dataset, along with satellite zenith angle, solar zenith angle, and geographic coordinates (latitude and longitude).  
To generate ground-truth labels, we use the 2B-CLDCLASS-LIDAR dataset, which provides \textit{base\_values}, \textit{top\_values}, and \textit{phase\_values}. Each pixel's vertical profile is divided into 38 layers from 0 to 19 km with 500 m resolution. Cloud top and bottom heights are derived from \textit{top\_values} and \textit{base\_values}, and cloud-type annotations (\textit{cloud\_mask}) are filled using \textit{phase\_values}, where 0 denotes no cloud, 1 ice cloud, 2 mixed cloud, and 3 liquid cloud.  
Finally, the matched AHI–CloudSat/Cloud Profiling Radar (CPR) contour points are segmented into spatial chunks for subsequent model training.
When the CloudSat radar passes over the target area, a matching AHI pixel is found within a ±5-minute time window. In spatial matching, LIDAR scan points are used to identify the nearest AHI pixel. Due to differing resolutions, multiple LIDAR points may correspond to a single AHI pixel. In such cases, discrete variables are weighted, and continuous variables are averaged. The final result is the LIDAR data corresponding to each AHI pixel.



\section{The Proposed Method}

As illustrated in Figure 2, the proposed network is a sparse-guided multi-scale attention generating network (\textit{SGMAGNet}), designed to generate high-quality 3D cloud phase classification maps. The architecture consists of four main components: spatial and height feature extraction, feature fusion, a multi-scale volumetric generator, and a phase-aware gating module. First, spatial and vertical features are extracted from the input AHI data. These features are then fused and progressively enhanced through the multi-scale generator. Finally, a phase-aware gating mechanism is employed to accurately classify the cloud phase state for each voxel in the 3D volume.

\begin{table*}[ht]
    \centering
    \caption{Performance Comparison of Models on Cloud Vertical Mask Classification. This table presents the performance metrics (Accuracy, Precision, Recall, F1 score, and IoU) of various models, such as, Sgmagnet, UNetCoNext, SegNet, AttnUNet, TransUNet, and MambaUNet on the cloud mask classification task.} 
    \begin{tabular}{cccccc}
        \toprule
        Models &Accuracy& Precision & Recall  & F1 & IoU  \\
        \midrule
        Sgmagnet& \textbf{0.922264} & 0.687362 & \textbf{0.858461} & \textbf{0.763442} & \textbf{0.617393} \\
        UnetCoNext& 0.908632 & 0.646183 & 0.82817 & 0.725944 & 0.569790 \\
        SegNet& 0.903583 & 0.640242 & 0.776438 & 0.701793 & 0.540586 \\
        AttnUNet&0.917694 & 0.672653 & 0.85073& 0.751286 & 0.601648 \\
        TransUNet& 0.912257 & \textbf{0.707085} & 0.682074 & 0.694355 & 0.531809 \\
        MambaUNet& 0.906363 & 0.648159 & 0.785658 & 0.710316 & 0.550767 \\
        \bottomrule
    \end{tabular}
\end{table*}

\subsection{Dynamic Height Encoder}
This module is responsible for extracting spatial and height features from the input data. Spatial features are extracted through multiple convolutional layers, while height features are processed using a learnable embedding layer. Specifically, spatial features are extracted via convolutional layers as follows:
\[
\mathbf{F}_{\mathrm{spatial}} = \mathrm{Conv2D}(\mathbf{X})
\]
where $\mathbf{X} \in \mathbb{R}^{B \times C \times H \times W}$ is the input data, and $\mathbf{F}_{\mathrm{spatial}} \in \mathbb{R}^{B \times E \times H \times W}$ represents the spatial features, with $B$ as the batch size, $C$ as the number of input channels, $H$ and $W$ as the height and width, and $E$ as the embedding dimension. The height features are generated using a learnable embedding layer:
\[
\mathbf{F}_{\mathrm{height}} = \mathrm{Repeat}(\mathbf{H}, B, W)
\]
where $\mathbf{H} \in \mathbb{R}^{1 \times H_{\dim} \times E}$ is the height embedding parameter, and $\mathbf{F}_{\mathrm{height}} \in \mathbb{R}^{B \times H_{\dim} \times W \times E}$ represents the height features.

\subsection{Multi Scale Volumetric Generator}
This module processes features at multiple scales (e.g., 1, 2, 4), enhancing the network's ability to learn across different resolutions. Features at each scale are processed using 3D convolutions, and the resulting features are upsampled to match the original size. The operation is described as follows:
\[
\mathbf{F}_{\mathrm{scaled}} = \mathrm{Conv3D}(\mathrm{Interp}(\mathbf{X}, \mathrm{scale}))
\]
where \textbf{Interp}($\mathbf{X}$, \texttt{scale}) represents downsampling or upsampling the input data $\mathbf{X}$, and $\mathbf{F}_{\mathrm{scaled}}$ is the feature processed at different scales. Features from multiple scales are then concatenated and fused:
\[
\mathbf{F}_{\mathrm{fused}} = \mathrm{Conv3D}(\mathrm{Concat}(\mathbf{F}_{\mathrm{scaled1}}, \mathbf{F}_{\mathrm{scaled2}}, \ldots, \mathbf{F}_{\mathrm{scaledn}}))
\]
The concatenated features are further processed using a 1x1 convolution, producing the final fused feature.

\subsection{Phase Aware Gating}
This module classifies the generated volumetric features into different phase states. The features are processed using 3D convolutions, followed by the application of a Softmax activation function to output the phase probabilities for each voxel. The operation is described by the following formula:
\[
\mathbf{F}_{\mathrm{phase}} = \mathrm{Softmax}(\mathrm{Conv3D}(\mathbf{X}))
\]
where $\mathbf{X}$ is the input feature, and $\mathbf{F}_{\mathrm{phase}} \in \mathbb{R}^{B \times N \times D \times H \times W}$ is the probability distribution over four phase classes, with $\mathbf{D}$ representing the depth, and $\mathbf{H}$ and $\mathbf{W}$ representing the height and width, respectively.

\section{Experiments and Results}
\subsection{Experimental Setup}
The model is implemented in Python using the PyTorch framework and trained on an NVIDIA RTX 4090 GPU. During training, the Adam optimizer with default settings is used. The input consists of a $16 \times 128 \times 128$ multi-channel image and auxiliary variables (e.g., latitude, longitude, zenith angles, and a nighttime flag). Sparse position masks are applied for loss calculation. The batch size is set to 4, and the learning rate is dynamically adjusted. In the testing phase, the input includes the same image and auxiliary information.
\subsubsection{Comparison Methods}
Five representative research results were selected for comparison, including UnetCoNext\cite{jeggle2024icecloudnet}, Segnet\cite{badrinarayanan2017segnet}, UnetAttn\cite{oktay2018attention}, TransUnet\cite{chen2021transunet}, MambaUnet\cite{wang2024mamba} , and the SGMAGNet model used in this study. An AHI-CloudSat/CALIPSO dataset, containing approximately 2800 multi-channel images, was used to evaluate the generalization ability of the method. The AHI-CC dataset has been divided into training, validation, and test sets in a ratio of 8:1:1. Cross-instrument matching and evaluation will be performed in future work.

\subsection{Comparison Results}
The task is to invert the vertical orientation of clouds at each pixel based on multi-channel imager data, which is essentially a multi-classification task involving 38 layers of vertical vectors. In numerical weather prediction, the determination of cloud presence is critical. To comprehensively assess model performance, two sets of weighted evaluation metrics are employed to address the uneven distribution of cloud existence and types. This approach effectively measures the model's ability to recognize cloud types that constitute a relatively small percentage of the total clouds and avoids the shortcomings of traditional metrics that may overlook these types.
\begin{table*}[ht]
    \centering
    \caption{Weighted Performance Evaluation of Cloud Type Classification Across Models. This table compares the weighted performance metrics (Balanced Accuracy, Kappa, Precision Macro, Recall Macro, and F1 Macro) of various models—Sgmagnet, UnetCoNext, SegNet, AttnUNet, TransUNet, and MambaUNet—on the cloud type classification task.} 
    \begin{tabular}{cccccc}
        \toprule
        Models &Balanced\_Accuracy&kappa & Precision\_Macro  & Recall\_Macro & F1\_Macro  \\
        \midrule
        Sgmagnet& \textbf{0.747} & \textbf{0.681} & \textbf{0.669} & \textbf{0.747} & \textbf{0.702} \\
        UnetCoNext& 0.686 & 0.641 & 0.636 & 0.686 & 0.646 \\
        Segnet& 0.647 & 0.613 & 0.624 & 0.647 & 0.622 \\
        AttnUNet& 0.726 & 0.674 & 0.652 & 0.726 & 0.682\\
        TransUnet& 0.568 & 0.605 & 0.541 & 0.568 & 0.554\\
        MambaUnet& 0.622 & 0.624 & 0.528 & 0.622 & 0.567 \\
        \bottomrule
    \end{tabular}
\end{table*}

\begin{figure*}[ht] 
\centering 
\includegraphics[scale=0.22]{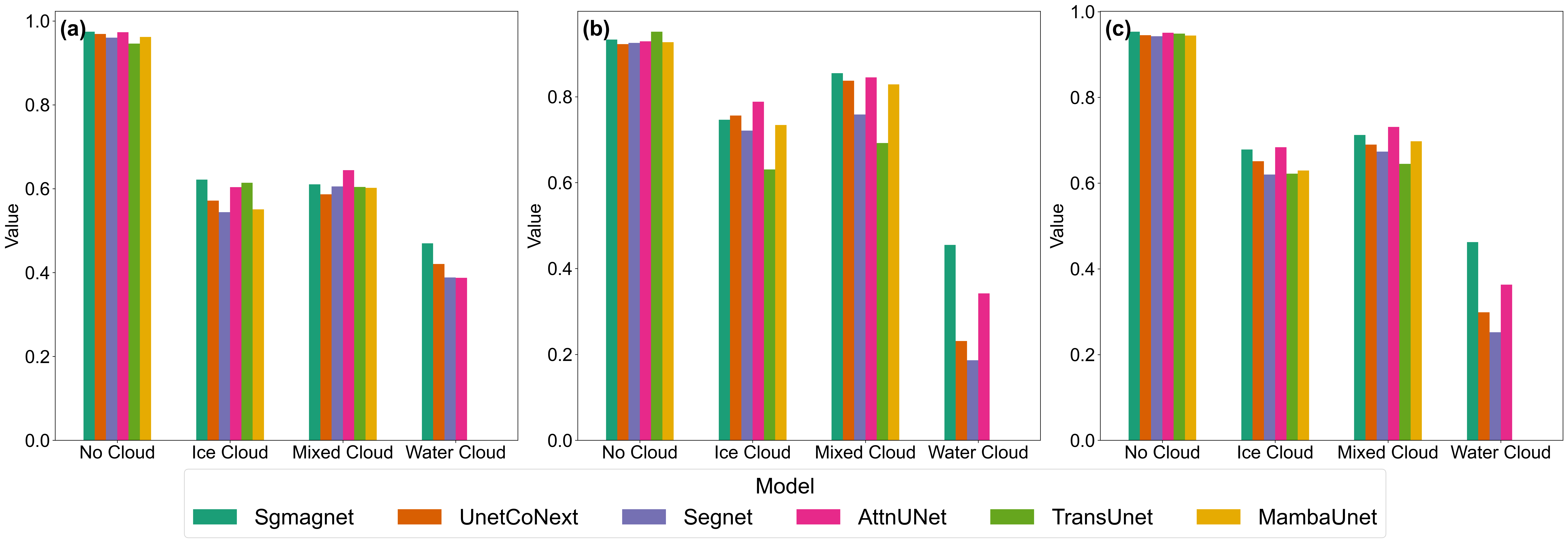} 
\caption{Macro-Average Precision, Recall, and F1-Score Across Cloud Types for Six Models. This figure compares the macro-average precision (a), recall (b), and F1-score (c) for six cloud phase prediction models across four cloud types (No Cloud, Ice Cloud, Mixed Cloud, Water Cloud), highlighting performance variations in cloud classification.} 
\label{fig1} 
\end{figure*}
\subsubsection{Cloud Mask Profile Evaluation.}
This evaluation metric is used to determine the presence or absence of clouds at each spatial location, i.e., cloud presence, Accuracy, Precision, Recall, F1 value (f1) and intersection and union ratio (IoU) which are formulated as follows.

$$\mathrm{Accuracy}=\frac{TP+TN}{TP+TN+FP+FN}$$
$$\mathrm{Precision}=\frac{TP}{TP+FP}$$
$$\mathrm{Recall}=\frac{TP}{TP+FN}$$
$$\mathrm{F1}=2\times\frac{\mathrm{Precision}\times\mathrm{Recall}}{\mathrm{Precision}+\mathrm{Recall}}$$
$$\mathrm{IoU}=\frac{|A\cap B|}{|A\cup B|}$$


Table 1 demonstrates the performance of \textit{Sgmagnet} and other models in the cloud presence determination task. \textit{Sgmagnet} achieves the best performance in terms of \textit{Precision} (0.922) and \textit{Recall} (0.858), with an \textit{F1} score of 0.763 and an \textit{IoU} of 0.617. In comparison, \textit{UnetCoNext} and \textit{AttnUNet} show similar performance, with \textit{Precision} and \textit{Recall} values of 0.646 and 0.673, and 0.828 and 0.851, respectively. \textit{SegNet} demonstrates more balanced results, though slightly lower in both \textit{Precision} and \textit{Recall}. \textit{TransUNet} yields higher \textit{Precision} but suffers from lower \textit{Recall}. \textit{MambaUNet} performs the worst across all metrics. Overall, \textit{Sgmagnet} outperforms the other models in the cloud presence classification task.
\subsubsection{Cloud Phase Profile Evaluation.}
This evaluation metric is used to assess the recognition correctness of each spatial location cloud type, including weighted accuracy (\textit{Balanced\_Accuracy}), Kappa coefficient (\textit{kappa}), macro precision (\textit{Precision\_Macro}), macro recall (\textit{Recall\_Macro}), and macro F1 value (\textit{F1\_Macro}). Their specific formulas are as follows, where $C$ denotes the number of cloud phase categories (e.g., no cloud, ice cloud, mixed cloud, liquid cloud).

$$\text{Balanced\_Accuracy}=\frac{1}{C}\sum_{i=1}^C\frac{TP_i}{TP_i+FN_i}$$
$$\mathrm{Kappa}=\frac{P_o-P_e}{1-P_e}$$
$$\text{Precision\_Macro}=\frac{1}{C}\sum_{i=1}^C\frac{TP_i}{TP_i+FP_i}$$
$$\text{Recall\_Macro}=\frac{1}{C}\sum_{i=1}^C\frac{TP_i}{TP_i+FN_i}$$
$$\mathrm{F1\_Macro}=\frac{1}{C}\sum_{i=1}^C\frac{2\times\mathrm{Precision}_i\times\mathrm{Recall}_i}{\mathrm{Precision}_i+\mathrm{Recall}_i}$$

Table 2 demonstrates that \textit{Sgmagnet} performs well in all the metrics, especially outperforming the other models in terms of \textit{Balanced\_Accuracy} (0.747) and \textit{F1\_Macro} (0.702). \textit{UnetCoNext} and \textit{AttnUNet} also perform better, particularly achieving scores close to \textit{Sgmagnet} in terms of \textit{Recall\_Macro} and \textit{F1\_Macro}. Overall, \textit{Sgmagnet} outperforms the other models in the cloud type recognition task, demonstrating strong classification ability and comprehensive performance.

\begin{figure*}[ht] 
\centering 
\includegraphics[scale=0.18]{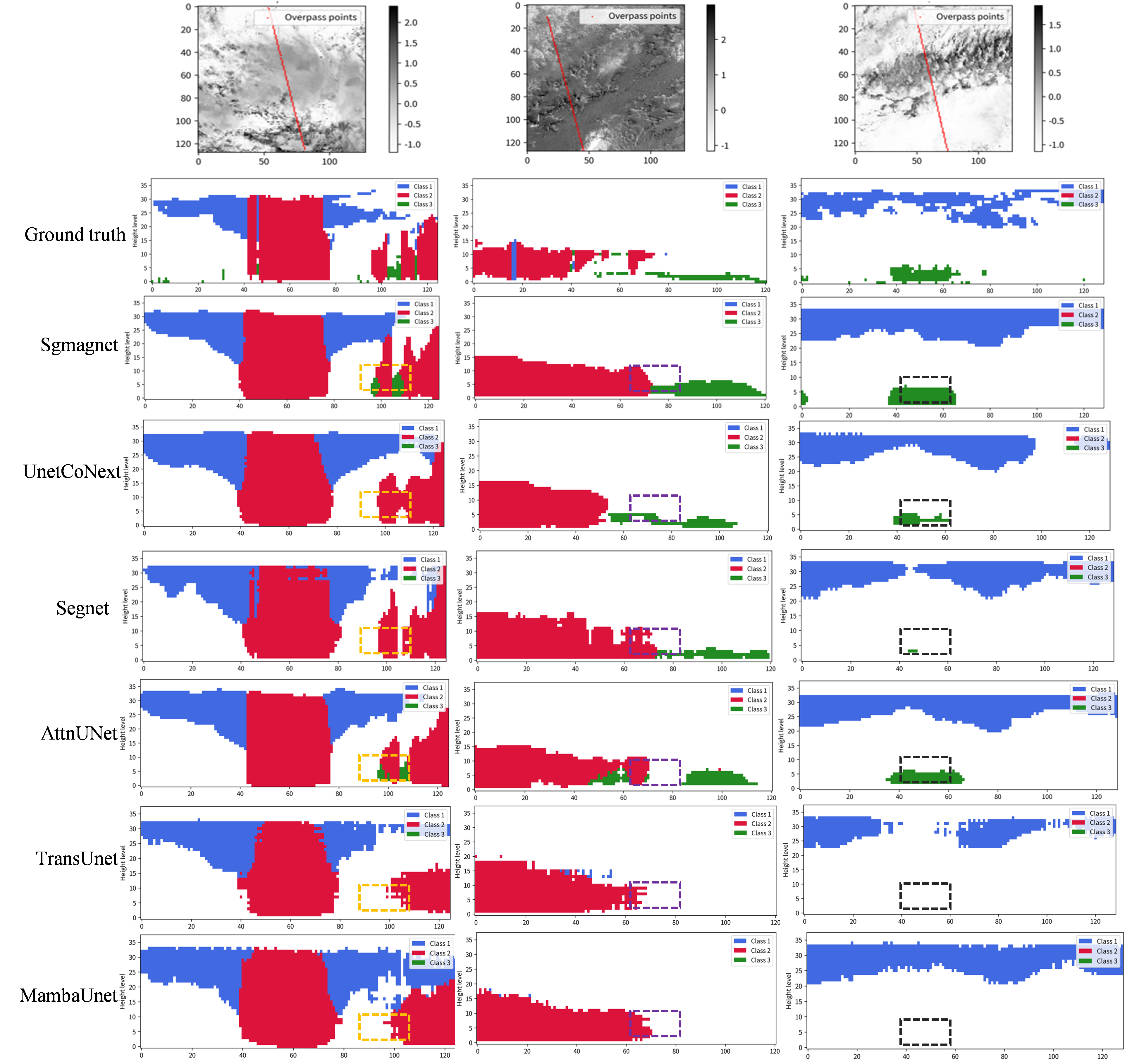} 
\caption{Visual Comparison of Cloud Phase Profile Predictions Along CALIOP-CPR Vertical Profiles. This figure compares cloud phase from models (Sgmagnet, UNetCoNext, SegNet, AttnUNet, TransUNet, MambaUNet) against ground truth along CALIOP-CPR vertical profiles. The orange, purple, and black boxes denote detailed cloud profile examples.} 
\label{fig1} 
\end{figure*}
As shown in Figure 3. Subfigures (a), (b), and (c) correspond to \textit{Precision}, \textit{Recall}, and \textit{F1-score}, respectively, across four cloud categories. \textit{SGMAGNet} demonstrates strong performance in the \textit{No Cloud}, \textit{Ice Cloud}, and \textit{Mixed Cloud} categories, achieving competitive scores compared to other models. Notably, it exhibits a significant advantage in recognizing \textit{Water Cloud}, outperforming all baselines by a large margin. In contrast, \textit{TransUNet} and \textit{MambaUNet} show almost no recognition ability for the \textit{Water Cloud} category, while maintaining relatively balanced performance on the remaining cloud types. These results highlight the superior capability of \textit{SGMAGNet} in handling difficult cases such as \textit{Water Cloud}, making it more effective for comprehensive cloud type classification.

Figure 4 shows analysis results for three AHI–CC matched cases with active sensor data. The first row (\textit{Patch Plane Visualization Map of AHI-8 with CALIOP-CPR Overpass}) depicts the spatial alignment of AHI-8 imagery and CALIOP-CPR tracks, with red points marking overpass locations. The second row (\textit{Vertical Cloud Phase Profile from CALIOP-CPR, denoted as CC}) presents the reference cloud structure, where white indicates no cloud, Class 1 (blue) is ice cloud, Class 2 (red) is mixed-phase, and Class 3 (green) is water cloud. The remaining rows show the predicted vertical cloud phase profiles from different models, including \textit{SGMAGNet}, \textit{UnetCoNext}, \textit{SegNet}, \textit{AttnUNet}, \textit{TransUNet}, and \textit{MambaUNet}. These visualizations enable a direct comparison of the models' ability to recover fine-scale vertical cloud structures along the CALIOP-CPR contours.
SGMAGNet excels in cloud classification, especially at the Class2 (mixed clouds) to Class3 (water clouds) transition, showing clear boundaries and fewer misclassifications (orange dashed box). Other models (UnetCoNext, Segnet, TransUnet, MambaUnet) struggle with complex regions and texture variations, while UnetAttn performs comparably but misclassifies at boundaries (purple dashed box). TransUnet and MambaUnet fail to identify Class3 effectively, aligning with poor statistical performance. SGMAGNet also excels at detecting water clouds under ice clouds (black dashed box). Overall, it accurately captures small changes and cloud hierarchy, minimizing misclassifications.

\section{ Conclusion}
This paper introduces a novel dataset for 3D cloud structure inversion, combining multi-channel passive imager data with CALIOP-CPR vertical profiles, annotated with voxel-level cloud phase labels. We propose \textit{SGMAGNet}, a network integrating dynamic encoding and a multi-scale volumetric generator, which outperforms baselines (e.g., \textit{UnetCoNext}, \textit{SegNet}, \textit{TransUNet}) in \textit{Precision}, \textit{F1-score}, and \textit{IoU}. \textit{SGMAGNet} excels at capturing class-boundary transitions and multi-layer structures, particularly distinguishing mixed-phase from liquid-phase clouds despite their similar radiative properties and dataset imbalance. By reducing misclassifications, it enhances generalization, offering a robust solution for cloud phase profiling.


\end{document}